\documentclass[letterpaper, 10 pt, conference]{IEEEtran}
\usepackage{amsmath,amsfonts}
\usepackage{algorithmic}
\usepackage{algorithm}
\usepackage{array}
\usepackage[caption=false,font=normalsize,labelfont=sf,textfont=sf]{subfig}
\usepackage{textcomp}
\usepackage{stfloats}
\usepackage{etoolbox}
\usepackage{url}
\usepackage{verbatim}
\usepackage{makecell}
\usepackage{graphicx}
\usepackage{booktabs}
\usepackage{enumitem}
\usepackage[colorlinks=true,citecolor=blue]{hyperref}
\begin{document}
\title{Language-Conditioned Representations and Mixture-of-Experts Policy for Robust Multi-Task Robotic Manipulation}

\author{\textit{Xiucheng Zhang}, \textit{Yang Jiang}, \textit{Hongwei Qin}, \textit{Jiashuo Bai}}

\markboth{Journal of \LaTeX\ Class Files,~Vol.~14, No.~8, August~2021}%
{Shell \MakeLowercase{\textit{et al.}}: A Sample Article Using IEEEtran.cls for IEEE Journals}


\maketitle

\begin{abstract}
Perceptual ambiguity and task conflict limit multi-task robotic manipulation via imitation learning. We propose a framework combining a Language-Conditioned Visual Representation (LCVR) module and a Language-conditioned Mixture-of-Experts Density Policy (LMoE-DP). LCVR resolves perceptual ambiguities by grounding visual features with language instructions, enabling differentiation between visually similar tasks. To mitigate task conflict, LMoE-DP uses a sparse expert architecture to specialize in distinct, multimodal action distributions, stabilized by gradient modulation. On real-robot benchmarks, LCVR boosts Action Chunking with Transformers (ACT) and Diffusion Policy (DP) success rates by 33.75\% and 25\%, respectively. The full framework achieves a 79\% average success, outperforming the advanced baseline by 21\%. Our work shows that combining semantic grounding and expert specialization enables robust, efficient multi-task manipulation.
\end{abstract}


\section{Introduction}
\IEEEPARstart{T}{he} pursuit of a generalist robotic agent, capable of performing a wide array of manipulation tasks~\cite{VLAzhonshu,rt1,rt2}, remains a significant open challenge. A primary obstacle lies in the inherent conflict that arises when a single policy network must learn a diverse suite of tasks. Even for conceptually dissimilar skills, competing gradient updates interfere with each other within the network's shared components~\cite{Gradientsurgery,famo,liu2021conflict}. This problem is substantially exacerbated by the ambiguity of multi-task settings: nearly identical visual scenes and gripper trajectories can correspond to entirely different user intents. In imitation learning, this high visual-motor similarity acts as a significant confounder, forcing the network to average contradictory actions. This "similar-input, different-output" dilemma is a key manifestation of the underlying model conflict~\cite{zhang2020causal}, causing performance to plateau at success rates far below what is required for real-world deployment.

While leading imitation learning baselines like DP~\cite{DP} and ACT~\cite{ACT} perform exceptionally well on single-task benchmarks~\cite{MBA,Hybrid,EquiBot,manicm,TactileAloha},
but struggle in multi-task scenarios due to ambiguity and gradient interference. While large Vision-Language-Action (VLA) models like RT-1,2~\cite{rt1,rt2}, $\pi_0$~\cite{pai0}, and OpenVLA~\cite{openvla} show promise, their large parameter counts and high latency are often impractical for real-time robotics, highlighting a need for a lightweight yet efficient model for multi-task manipulation.

To address the limitations of existing multi-task imitation learning methods, we propose a new framework. It consists of two key components: the Language-Conditioned Visual Representation (LCVR) module and the Language-conditioned Mixture-of-Experts Density Policy (LMoE-DP).
LCVR resolves the ''similar-input, different-output'' ambiguity by grounding visual features with language instructions, achieving 33.75\% and 25\% improvements over the baseline methods ACT~\cite{ACT} and DP~\cite{DP}, respectively. Building on this, the LMoE-DP policy enhances task specialization and optimization stability by dedicating specialized experts to distinct, multimodal action distributions and modulating gradients to mitigate destructive interference. On a challenging five-task real-robot benchmark, LMoE-DP achieves 79\% average success, outperforming the baseline $\Sigma$-agent~\cite{agent} by 21\%. These results demonstrate the effectiveness of semantic grounding and expert specialization in advancing multi-task robotic manipulation.
The main contributions of this work are summarized as follows:
\begin{itemize}[leftmargin=*, topsep=0pt, itemsep=-2pt, parsep=0pt]
  \item We design the LCVR module, a lightweight and plug-and-play visual encoder that aligns high-resolution visual features with task instructions. This module efficiently resolves perceptual ambiguity by grounding visual features with semantic cues, enabling multi-task manipulation with minimal computational overhead.
  \item We propose the LMoE-DP, which uses a sparse MoE architecture to mitigate task conflict. Its specialized Mixture Density Network (MDN) experts capture complex multimodal actions, while integrated gradient modulation ensures stable training and robust multi-task performance.
\end{itemize}

\section{Related Works}
\subsection{Imitation learning for robotic manipulation}
Imitation learning provides an effective paradigm for robots to acquire manipulation skills directly from demonstrations. Classical Behavioral Cloning (BC)~\cite{BC} methods offer simplicity but are limited by distributional shift, which often causes compounding errors~\cite{DAgger}. To address this, recent approaches incorporate more expressive architectures. Transformer-based models such as ACT leverage sequence modeling to capture temporal dependencies, while methods like Implicit Behavioral Cloning (IBC)~\cite{IBC} and diffusion-based policies (DP~\cite{DP}, DP3~\cite{DP3}) better capture multimodal action distributions through energy-based modeling and stochastic denoising processes. These advancements have significantly improved performance on single-task benchmarks and demonstrated the scalability of imitation learning to increasingly complex skills. However, in multi-task settings, aforementioned challenges remain when diverse tasks share high visual-motor similarity, which can hinder task specialization.

\subsection{Multi-Task learning}
A key challenge in multi-task manipulation is resolving ambiguity when visual scenes are similar across different tasks.This ambiguity can be addressed through various methods, including leveraging multi-modal information, fine-grained descriptions, contextual awareness, and advanced techniques such as world models~\cite{surfer}. To this end, recent methods focus on learning more discriminative representations. One dominant strategy is to leverage 3D vision to better capture scene geometry; approaches like C2F-ARM~\cite{c2f}, PerAct~\cite{peract}, 3D Diffuser Actor~\cite{3ddiffuser}, RVT~\cite{rvt1} and RVT-2~\cite{rvt2} process inputs as voxels or point clouds to disambiguate actions spatially. Another strategy, employed by methods such as the $\Sigma$-agent~\cite{agent}, explicitly aligns visual features with language instructions through contrastive learning, forcing the model to learn the semantic differences between tasks.Nonetheless, these strategies are often constrained by high computational costs or struggle to ground language in specific, task-relevant visual cues.

\section{Method}
\begin{figure}[htbp]       
  \centering
  \includegraphics[width=\linewidth]{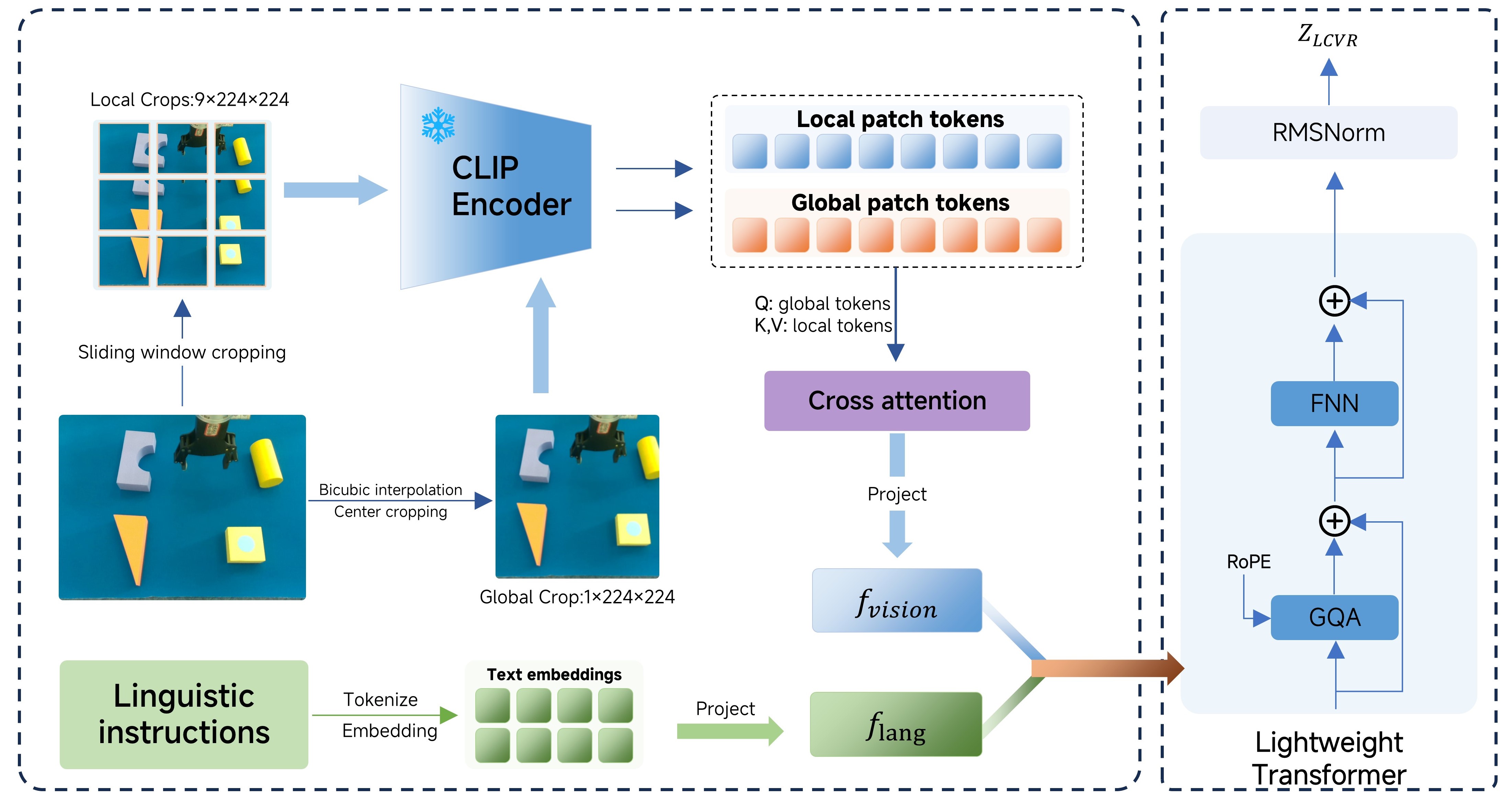}  
  \caption{The LCVR module architecture. From a high-resolution image, nine local and one global patch are extracted and then encoded by a shared, general-purpose pre-trained and frozen CLIP ViT-B/16(\href{https://huggingface.co/openai/clip-vit-base-patch16}{Clip-vit-base-patch16}). The resulting patch tokens are fused via cross-attention (global as Q, local as K/V) to synthesize a unified visual feature. This feature is then concatenated with an embedding from the language instruction and processed by a lightweight Transformer to produce the final language-conditioned representation, $z_{\text{LCVR}}$, for the downstream LMoE-DP policy.}
  \label{lcvr}
\end{figure}

\begin{figure*}[t]   
  \centering
  \includegraphics[width=\linewidth]{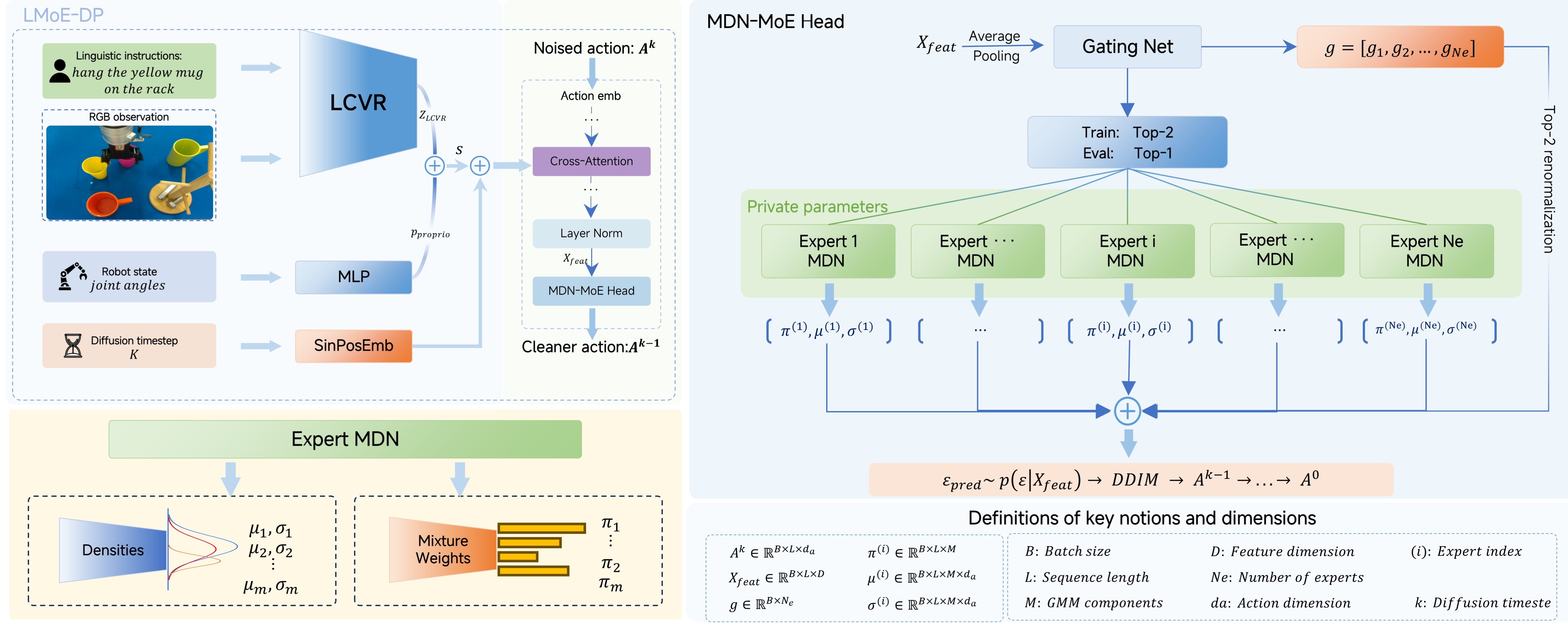}
  \caption{An overview of the LMoE-DP architecture. The policy is a conditional diffusion model. A Transformer backbone, conditioned on the language-visual feature $z_{\text{LCVR}}$, robot state, and diffusion timestep $k$, processes the noisy action sequence $A^k$ to produce a feature representation $X_{\text{feat}}$. A sequence-level gating network routes this representation to a bank of specialized MDN experts using a Top-2 strategy during training and a Top-1 strategy for inference. Each active MDN expert parameterizes a Gaussian Mixture Model (GMM) to predict the noise $\epsilon_{\text{pred}}$, which is used by the DDIM scheduler to iteratively compute the cleaner action $A^{k-1}$.}
  \label{LMOEDP}
\end{figure*}

Our framework consists of three components: the LCVR module (Fig.~\ref{lcvr}) for semantically-grounded state representations, the LMoE-DP policy (Fig.~\ref{LMOEDP}) with a sparse MoE architecture to address task conflict, and gradient modulation via the FAMO algorithm to stabilize training and ensure effective expert specialization.

\subsection{Language-Conditioned Visual Representation (LCVR)}
LCVR produces a semantically grounded and spatially accurate state representation by preserving fine-grained details often lost in downsampling.

LCVR uses a multi-scale encoding strategy to preserve information. From a high-resolution image $I \in \mathbb{R}^{480 \times 640 \times 3}$, we extract one global feature ($f_{\text{global}}$) and nine local patches ($\{f_{\text{local}}^i\}_{i=1}^9$) using a frozen, pre-trained CLIP encoder~\cite{CLIP}. This captures both holistic context and high-fidelity details without the information loss of naive resizing~\cite{vit}. The features are unified into a single representation ($f_{\text{vision}}$) via cross-attention~\cite{attention}, where the global feature (Query) attends to the local features (Keys/Values), focusing on task-relevant details.

The synthesized visual features are then conditioned with linguistic instructions to resolve any remaining task ambiguity. The textual command is tokenized and projected into an embedding sequence, $f_{\text{lang}}$, of a compatible dimension. This sequence is then concatenated with the visual representation $f_{\text{vision}}$. The final, deeply integrated representation is produced by a lightweight Transformer encoder optimized for real-time robotics, which incorporates Grouped-Query Attention (GQA)~\cite{GQA}, RMSNorm~\cite{rmsnorm}, and Rotary Position Embedding (RoPE)~\cite{rope} to improve computational efficiency and training stability. The final output of the module, $z_{\text{LCVR}}$, is a dense, disambiguated, language-conditioned state representation.

\subsection{Language-conditioned Mixture-of-Experts Density Policy (LMoE-DP)}
While LCVR addresses perceptual ambiguity at the input level, task conflict can still arise within the policy network. To address this, we introduce the LMoE-DP to promote task specialization and explicitly model multimodal action distributions.

The denoising network is conditioned on a comprehensive state representation ($s$) formed by concatenating two distinct sources of information: the language-conditioned feature $z_{\text{LCVR}}$ from our LCVR module and the robot's proprioceptive state $p_{\text{proprio}}$ (e.g., joint angles). This fusion provides the policy with a multi-faceted understanding of the task, combining high-level semantic context and the robot's physical state.

We replace the monolithic regression head with a sparse MoE architecture~\cite{moe}, in which each expert is implemented as a MDN~\cite{MDN}. This design allows the model to dedicate specialized experts to learn the unique data distributions of different tasks, enabling it to effectively model task-specific action distributions.

The core of our approach is a denoising network with a Transformer-based architecture, which we denote as a whole by $F_T$. At each step of the reverse diffusion process, this network takes the noisy action trajectory $A^k$, the diffusion timestep $k$, and the comprehensive conditioning observation $s$ as input. After its internal encoder-decoder process, it produces a final sequence of rich feature representations, $X_{\text{feat}}$:
\begin{equation}
X_{\text{feat}} = F_T(A^k, k, s)
\end{equation}
where $X_{\text{feat}} \in \mathbb{R}^{L \times D}$, with $L$ being the action sequence length and $D$ the feature dimension. This feature sequence then serves as the input to our specialized MoE output layer, which operates as follows:

\textbf{Sequence-Level Gating with Sparse Routing.} To ensure that generated action trajectories are coherent, we employ a sequence-level gating mechanism. Instead of making a routing decision for each token, the Gating Network $G$ makes a single decision for the entire sequence. It first aggregates the token-level features in $X_{\text{feat}}$ into a single sequence feature $x_{\text{seq}} \in \mathbb{R}^{D}$ (via average pooling). This feature is then used to compute routing probabilities $g$ for the $N_e$ experts:
\begin{equation}
g = \text{Softmax}(x_{\text{seq}} W_g) \in \mathbb{R}^{N_e}
\end{equation}
where $W_g$ are the weights of the gating network. We employ a sparse routing strategy. During \textbf{training}, the total loss is calculated based on the outputs of the 2 experts with the highest gating probabilities (soft top-2). For \textbf{inference}, we select the single expert, $i^*$, with the highest gating probability ($i^* = \text{argmax}_i(g_i)$), which guarantees that the entire trajectory is generated by one consistent policy (hard top-1).

\textbf{Mixture Density Prediction and Parameterization.} Each parallel MDN expert $E_i$ takes the feature sequence $X_{\text{feat}} \in \mathbb{R}^{L \times D}$ and outputs a sequence of GMM parameters for all timesteps at once: $\{\pi_{m,t}^{(i)}, \mu_{m,t}^{(i)}, \sigma_{m,t}^{(i)}\}_{t=1}^{L}$, where each timestep $t$ corresponds to a Gaussian Mixture Model with $M$ components. All timesteps are processed in parallel without autoregression. Specifically, the expert network parameterizes the GMM as follows:
\begin{itemize}[leftmargin=*, topsep=0pt, itemsep=-2pt, parsep=0pt]
  \item Mixing Coefficients: It outputs logits $\alpha_{m,t}^{(i)}$ which are transformed via a Softmax function to produce the normalized weights $\pi_{m,t}^{(i)}$.
  \item Means: It directly outputs the mean vector $\mu_{m,t}^{(i)}$ for each component.
  \item Standard Deviations: It outputs log standard deviations $\hat{\sigma}_{m,t}^{(i)}$, which are then transformed using $\sigma_{m,t}^{(i)} = \exp(\hat{\sigma}_{m,t}^{(i)}) + \sigma_{\min}$, where $\sigma_{\min}$ is a small constant (e.g., $10^{-4}$) to ensure positivity and prevent mode collapse.
\end{itemize}

During \textbf{training}, we model the joint distribution of the full target noise sequence $\epsilon = (\epsilon_1, \dots, \epsilon_L)$. The predicted probability density is a weighted combination from the top-2 experts (denoted by the set $S$):
\begin{equation}
\resizebox{.9\linewidth}{!}{%
$\displaystyle
p(\epsilon \mid X_{\text{feat}}) = \sum_{i \in S} g'_i \prod_{t=1}^L \left( \sum_{m=1}^{M} \pi_{m,t}^{(i)} \mathcal{N}\!\bigl(\epsilon_t \mid \mu_{m,t}^{(i)}, \text{diag}(\sigma_{m,t}^{(i)})^2\bigr) \right)
$}
\end{equation}
where $g'_i = g_i / \sum_{j \in S} g_j$ are the renormalized gating weights. The training loss is the negative log-likelihood of this joint density:
\begin{equation}
\scalebox{0.8}{%
$
\displaystyle
\mathcal{L}_{\text{MDN}}
= -\log\!\sum_{i\in S} g'_i
\exp\!\Bigl(
      \sum_{t=1}^{L}
      \log\sum_{m=1}^{M}
      \pi_{m,t}^{(i)}\,
      \mathcal{N}\!\bigl(\epsilon_t \bigm|
                        \mu_{m,t}^{(i)},
                        \operatorname{diag}\!\bigl(\sigma_{m,t}^{(i)2}\bigr)
                        \bigr)
    \Bigr).
$}
\end{equation}In practice, this loss is computed using the log-sum-exp trick over both timesteps and mixture components to ensure numerical stability.

For \textbf{inference}, we first select the single most appropriate expert $i^*$ with the highest gating probability once for the entire sequence. Then, for each timestep $t$, we choose the GMM component $m^*$ \textbf{within this fixed expert} with the highest mixing coefficient: $m^* = \arg\max_m \pi_{m,t}^{(i^*)}$. The mean of this component, $\mu_{m^*,t}^{(i^*)}$, is used as the predicted noise $\epsilon_{\text{pred},t}$. The full predicted sequence $\epsilon_{\text{pred}} = (\epsilon_{\text{pred},1}, \dots, \epsilon_{\text{pred},L})$ is then passed to the DDIM scheduler to compute a denoised action sequence $A^{k-1}$, and this process is iterated until the final clean action sequence $A^0 = (a_t, a_{t+1}, \dots, a_{t+L-1})$ is obtained.

To ensure stable training and balanced expert utilization, we introduce a weighted auxiliary loss, $\mathcal{L}_{\text{aux}}$, to the main training objective~\cite{largemoe}:
\begin{equation}
\mathcal{L}_{\text{aux}} = \alpha \mathcal{L}_{\text{load}} + \beta \mathcal{L}_{\text{importance}}
\end{equation}
This loss combines a load balancing term ($\mathcal{L}_{\text{load}}$), which encourages routing an equal number of examples to each expert, and an importance term ($\mathcal{L}_{\text{importance}}$), which promotes balanced routing weights across a batch. Together, these losses prevent issues like expert collapse and encourage a stable division of labor.

\subsection{Stabilizing Expert Training with Gradient Modulation}
Although the MoE structure encourages specialization, the shared network backbone remains vulnerable to conflicting gradients from different experts---a phenomenon known as negative transfer~\cite{standley2020tasks,Gradientsurgery,famo,ruder2017overview}. This problem is particularly pronounced in sparse MoE architectures, where experts often model highly diverse task distributions. To mitigate this, we modulate updates to all shared parameters $\theta_{\text{shared}}$ using Fast Adaptive Multi-task Optimization (FAMO)~\cite{famo}.

At each training step, we first compute the gradient $g_i = \nabla_{\theta_{\text{shared}}} \mathcal{L}_i$ for each of the $k$ active experts, where $\mathcal{L}_i$ denotes the total loss contributed by expert $i$ (including MDN negative log-likelihood and auxiliary terms). FAMO then computes a set of non-negative coefficients $\{\alpha_i\}$ by solving a constrained optimization problem that minimizes the $\ell_2$ norm of the combined gradient while ensuring it remains a descent direction for each expert. This optimization admits a fast, closed-form solution based on the min-norm principle~\cite{desideri2012multiple}, and the resulting coefficients are normalized such that $\sum_{i=1}^k \alpha_i = 1$. The modulated gradient is then given by:
\begin{equation}
g_{\text{FAMO}} = \sum_{i=1}^{k} \alpha_i g_i
\end{equation}
which is used to update the shared parameters.

In addition, to ensure that the gating network continues to receive task-relevant learning signals, we supplement the FAMO-modulated gradient with the gradient of the mixture-level negative log-likelihood $\nabla_{\theta_{\text{shared}}}\mathcal{L}_{\text{mix}}$. The final update applied to shared parameters is therefore
\begin{equation}
g_{\text{shared}} = g_{\text{FAMO}} + \nabla_{\theta_{\text{shared}}}\mathcal{L}_{\text{mix}} .
\end{equation}

This ensures that updates follow a Pareto-efficient descent direction that minimizes gradient conflict while still encouraging effective routing in the gating network, thereby preventing destructive interference~\cite{sener2018multi}. As a result, expert specialization becomes more stable, and the shared representation converges more effectively.

\section{experiments}
\subsection{Experiment Setup}
\begin{figure}[htbp]
\centering
\includegraphics[width=\linewidth,totalheight=0.7\linewidth,keepaspectratio]{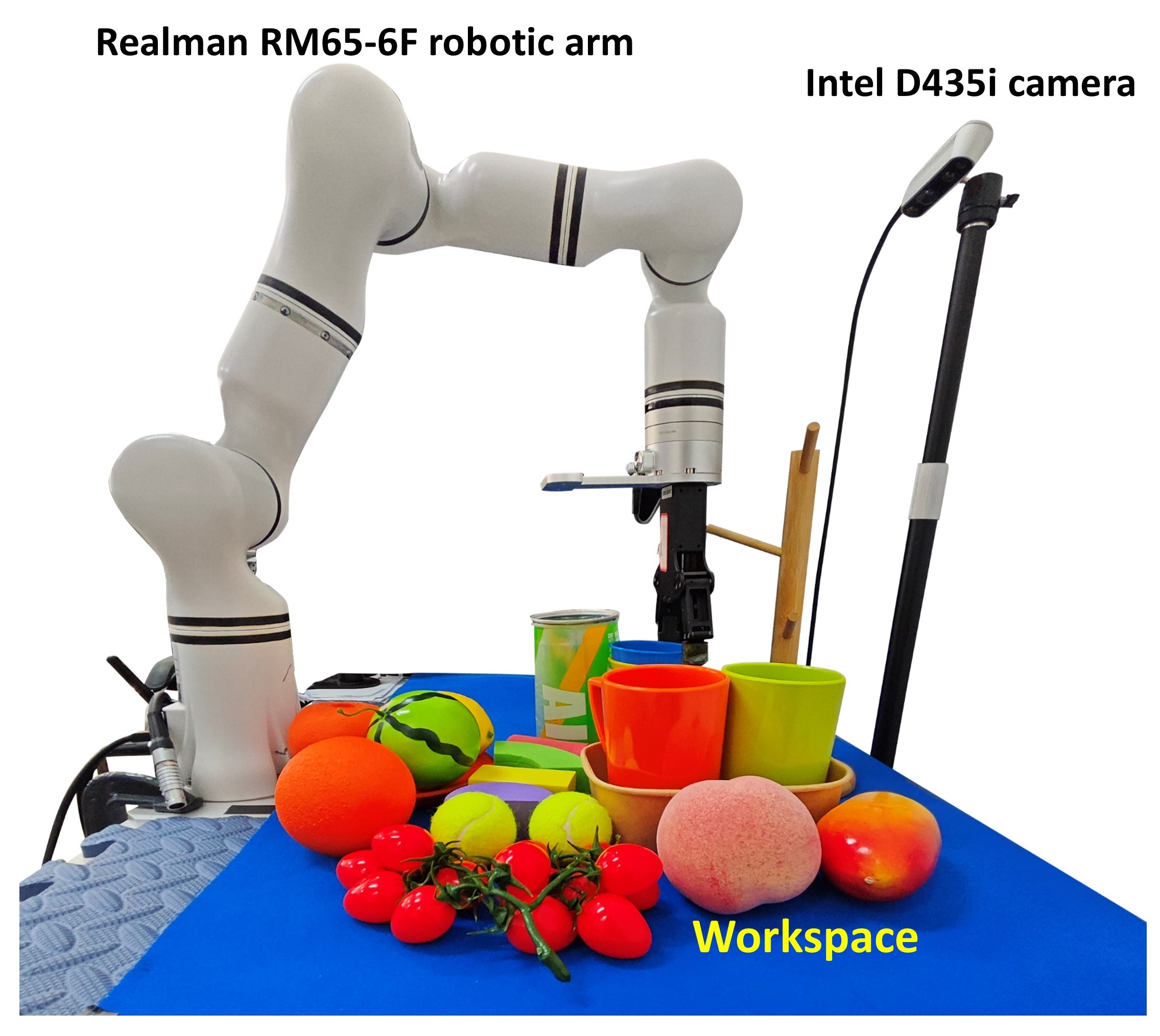}
\caption{Real-World Experimental Platform}
\label{setup}
\end{figure}
\textbf{Experimental Objectives:} This evaluation addresses three research questions: (Q1) Can the LCVR module leverage language to resolve perceptual ambiguity and improve multi-task performance? (Q2) Does the LMoE-DP architecture mitigate task conflict and model complex action distributions more effectively than a monolithic policy? (Q3) Does gradient modulation (FAMO) stabilize training, leading to improved expert specialization and higher task success rates?

\textbf{Platform:} Our experimental testbed (Fig.\ref{setup}) consists of a 6-DoF Realman robotic arm equipped with an EG2-4C2 gripper and an overhead Intel RealSense D435i camera for visual perception. All models were trained on a single NVIDIA H100 GPU (~3 hours per run), with inference performance evaluated on an NVIDIA 4060 Ti (16GB).

\textbf{Protocol:} We collected a dataset of 50 expert demonstrations for each task or experimental variant via human teleoperation. To promote policy generalization and prevent overfitting, the initial positions and orientations of all objects in the workspace were subjected to minor random perturbations in each demonstration. For evaluation, a trial is considered successful if the robot grasps the correct object and places it in the goal region within a 60s timeout. No human intervention is permitted. A failure is recorded if the robot fails to grasp, drops the object, or places it outside the designated area. The final performance is measured by the success rate, calculated as the fraction of successful trials.

\subsection{Effectiveness of LCVR}
\begin{figure}[htbp]
  \centering
  \includegraphics[width=\linewidth]{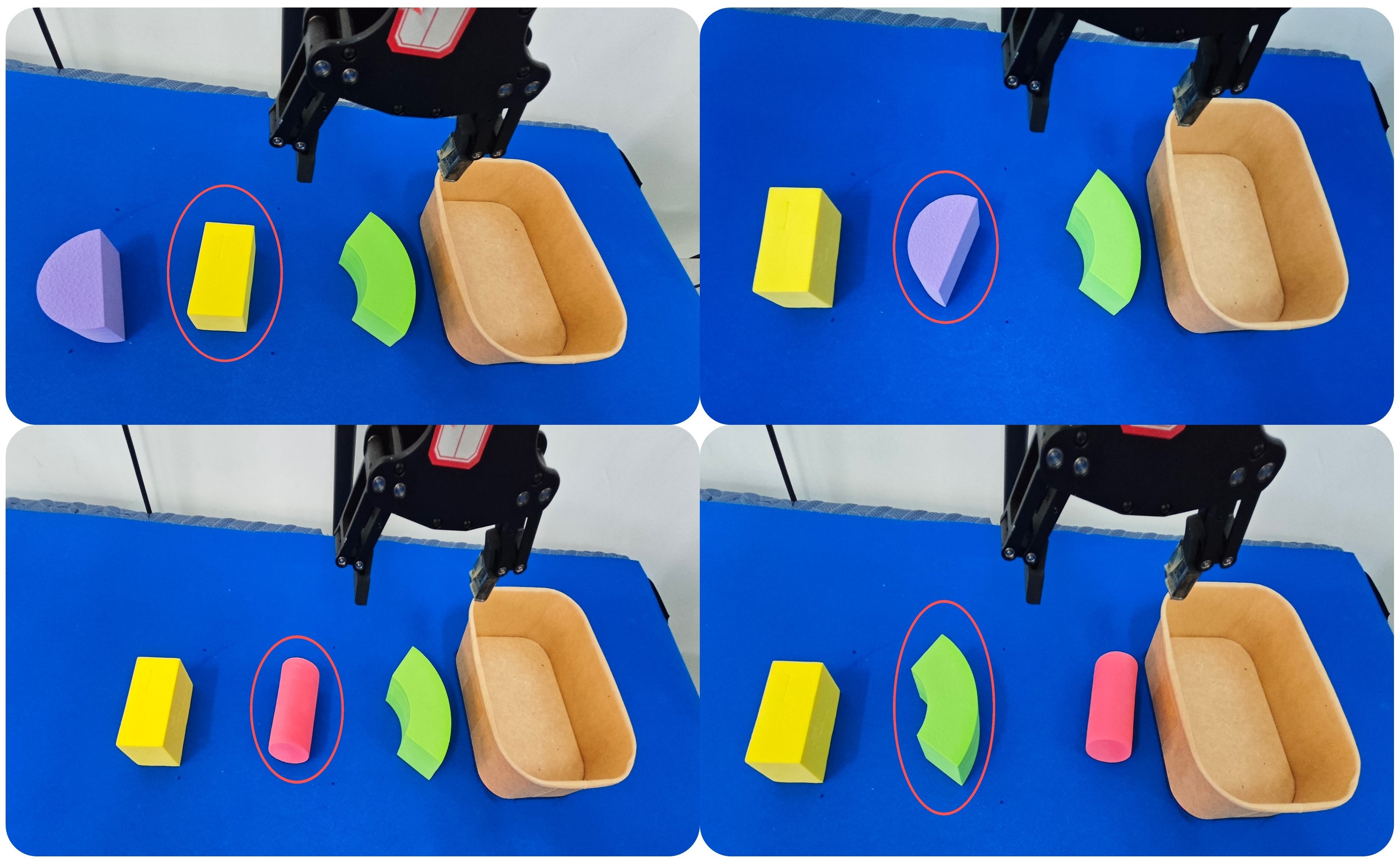}
  \caption{The four manipulation tasks used for evaluation. Target objects are indicated by red circles.}
  \label{LCVRexperiment}
  \end{figure}
\label{sec:lcvr_effectiveness}
To evaluate LCVR's ability to resolve the ``similar-input, different-output'' ambiguity, we designed a benchmark with four manipulation tasks (Fig.~\ref{LCVRexperiment}). To ensure a fair comparison with baselines that lack language input, we introduced a distinct positional prior: the target object was always placed centrally between two distractors, providing a key non-linguistic cue. We then integrated LCVR into the ACT~\cite{ACT} and DP~\cite{DP} pipelines by replacing their standard ResNet-18 visual encoder. For each method, a single policy was trained jointly on all four tasks and evaluated over 20 independent trials per task.

Table~\ref{tab:lcvr_effectiveness} reports the per-task success rates, which demonstrate the effectiveness of language conditioning over simple positional cues. Integrating LCVR led to significant improvements in performance across the board, with the average success rates for ACT and DP rising from 26.25\% and 32.50\% to 60\% and 57.5\%, respectively. This indicates that while the central positional prior provides a useful signal, LCVR's semantic grounding is significantly more effective and robust for resolving visual ambiguity. The improvements were observed across all four tasks, with the largest gains on tasks requiring fine-grained differentiation where baselines struggled (e.g., yellow rectangle and green arch). Even for tasks where the baseline already performed reasonably well, such as the red cylinder, LCVR still provided a notable improvement. Overall, these results provide a direct and affirmative answer to our first research question (Q1), confirming that the LCVR module effectively mitigates perceptual ambiguity by leveraging both its fine-grained visual representation and language conditioning, thereby substantially improving multi-task manipulation performance compared to policies that rely on standard visual encoders or simple environmental priors.
\begin{table}[t] 
\centering
\caption{Success rates (\%) on the high-visual-similarity benchmark.}
\label{tab:lcvr_effectiveness}
\renewcommand{\arraystretch}{1.4} 
\begin{tabular*}{\columnwidth}{@{\extracolsep{\fill}}cccccc}
\toprule[0.8pt]
\textbf{Method} & \makecell{Yellow \\ Rectangle} & \makecell{Purple \\ Semi-circle} & \makecell{Red \\ Cylinder} & \makecell{Green \\ Arch bridge} & \textbf{Average}\\
\midrule
ACT~\cite{ACT}             &    20  &   25    &   45    &    15   &    26.25   \\
\makecell{ACT with \\ LCVR} &   \textbf{65}    &   55    &   \textbf{70}    &   50    &    \textbf{60.00}  \\
\midrule
DP~\cite{DP}              &    10   &   35    &   55    &   30    &    32.50   \\
\makecell{DP with \\ LCVR}  &    50   &   \textbf{60}    &   65    &   \textbf{55}    &   57.50   \\
\bottomrule[0.8pt]
\end{tabular*}
\end{table}
  
\subsection{Evaluating task specialization with LMoE-DP}
\begin{figure}[htbp]
  \centering
  \includegraphics[width=\linewidth]{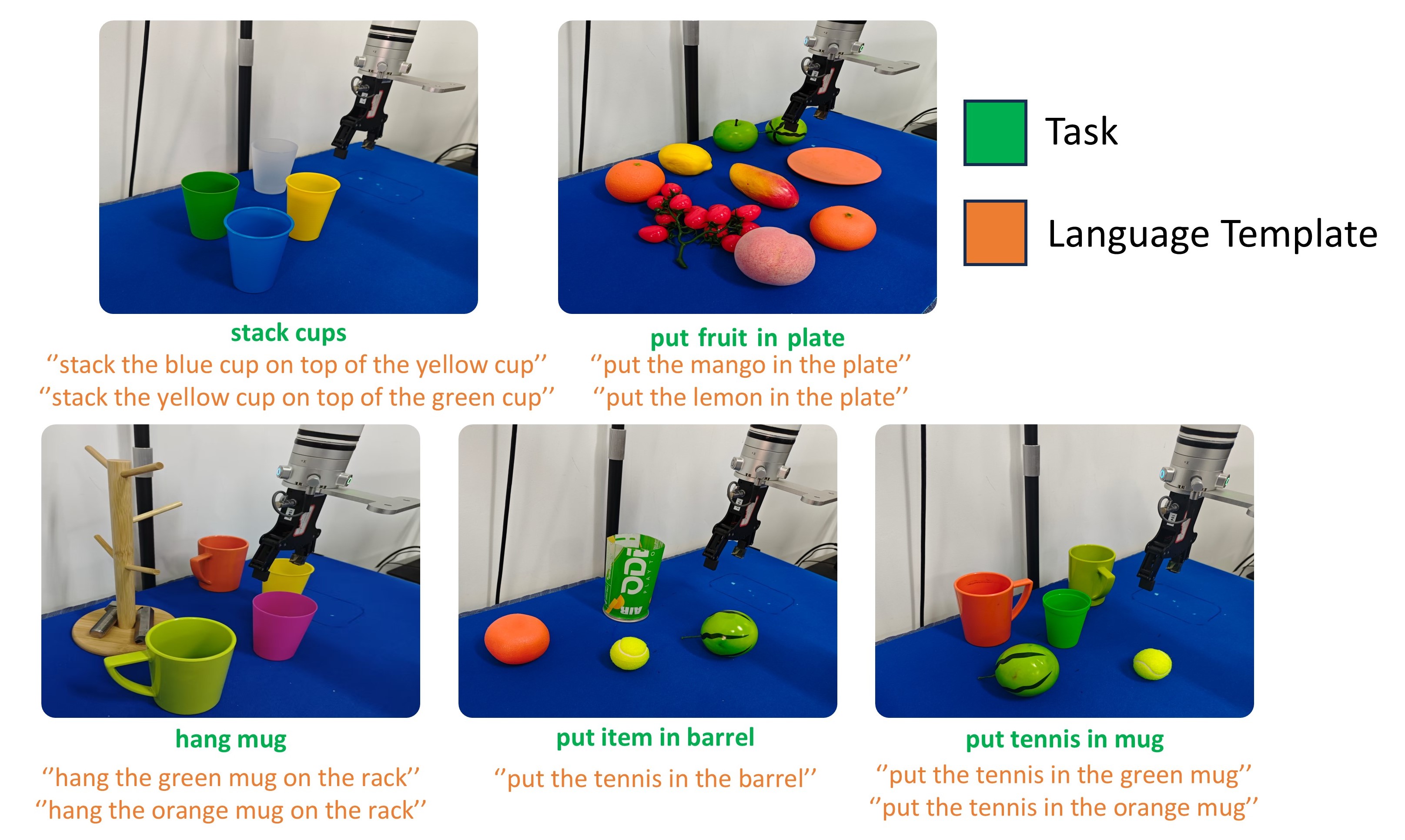}
  \caption{Illustration of the 5 tasks and their 9 variants, including the specific language templates used for each task.}
  \label{Nine}
  \end{figure}
To evaluate whether LMoE-DP facilitates expert specialization and alleviates task conflict, we compared it against the state-of-the-art $\Sigma$-agent~\cite{agent}, which explicitly aligns vision and language through contrastive objectives. Both approaches were trained and evaluated on the same multi-task dataset to ensure fairness. In our default configuration, the LMoE-DP head employs $N_e=4$ experts and $M=5$ Gaussian components per expert. These values were chosen based on a hyperparameter grid search, detailed below, which was designed to find an effective balance between task specialization and per-expert modeling capacity.
 
We adopt the five representative manipulation tasks from the $\Sigma$-agent benchmark~\cite{agent}, which capture different sources of ambiguity and conflict (Fig.~\ref{Nine}). Specifically, stack cups and put tennis in mug involve visually similar objects, where language disambiguation is crucial; put fruit in plate introduces ambiguity in spatial placement; while hang mug and put item in barrel differ substantially in their underlying action distributions compared to the others (e.g., lifting and hooking versus dropping and inserting). Together, these tasks cover the challenging scenario where similar inputs may correspond to distinct actions, and where multimodal action distributions must be captured for reliable execution.

Table~\ref{tab:lmoedp_specialization} summarizes the results, showing that LMoE-DP significantly outperforms the \(\Sigma\)-agent baseline. Our method achieves an average success rate of 79\%, a substantial +21\% improvement over the \(\Sigma\)-agent's 58\%. The most pronounced gains are on tasks with highly distinct action distributions, which often cause negative transfer in monolithic policies. For example, LMoE-DP achieves +30\% and +35\% improvements on ``hang mug'' (hooking motion) and ``put item in barrel'' (inserting motion), respectively. This strongly suggests the MoE architecture successfully assigns these unique skills to specialized experts, isolating their conflicting gradient updates from other tasks. Likewise, on tasks like ``stack cups'' (+25\%) and ``put tennis in mug'' (+30\%), where precise manipulation and semantic disambiguation are critical, LMoE-DP excels by routing semantically-grounded representations from LCVR to the appropriate expert. While the \(\Sigma\)-agent was more effective on the ``put fruit in plate'' task, which primarily involves spatial ambiguity, LMoE-DP's strong overall performance confirms its ability to mitigate task conflict and model multimodal actions.
  
\begin{table*}[t]
\centering
\caption{Per-task success rates (\%) over five multi-task categories. The evaluation covers 9 distinct task variants, each tested 10 times.}
\label{tab:lmoedp_specialization}
\renewcommand{\arraystretch}{1.4}
\begin{tabular*}{0.9\textwidth}{@{\extracolsep{\fill}}lccccc c}
\toprule[0.8pt]
\textbf{Method} & Stack cups & Put fruit in plate & Hang mug & Put item in barrel & Put tennis in mug & \textbf{Average} \\
\midrule
$\Sigma$-agent~\cite{agent} & 50 & \textbf{85} & 60 & 40 & 55 & 58 \\
LMoE-DP (Ours)              & \textbf{75} & 70 & \textbf{90} & \textbf{75} & \textbf{85} & \textbf{79} \\
\bottomrule[0.8pt]
\end{tabular*}
\end{table*}

\begin{table}[t]
\centering
\caption{Success rate (\%) grid search over number of experts ($N_e$) and MDN components ($M$).}
\label{tab:grid_success}
\renewcommand{\arraystretch}{1.5}
\begin{tabular*}{\columnwidth}{c @{\extracolsep{\fill}} ccccc|c}
\toprule
$N_e \backslash M$ & 1 & 3 & 5 & 7 & 9 & \textbf{Avg.} \\
\midrule
1 & 0 & 12 & 28 & 24 & 0 & 12.8 \\
2 & 8 & 48 & 52 & 44 & 8 & 32.0 \\
3 & 20 & 64 & 68 & 52 & 12 & 43.2 \\
4 & 12 & 72 & \textbf{92} & 68 & 4 & \textbf{49.6} \\
5 & 18 & 65 & 72 & 56 & 0 & 42.2 \\
\midrule
\textbf{Avg.} & 11.6 & 52.2 & \textbf{62.4} & 48.8 & 4.8 & \\
\bottomrule
\end{tabular*}
\end{table}

To choose an appropriate trade-off between expert count and per-expert MDN capacity, and to study their interaction, we ran a controlled grid search varying the number of experts $N_e\in\{1,2,3,4,5\}$ and the number of Gaussian components $M\in\{1,3,5,7,9\}$. For each cell in the grid we evaluated the trained policy on five representative tasks (one canonical variant per task) and performed five independent inference trials per variant, yielding $5\times 5=25$ runs per grid cell. The aggregated success rates for each $(N_e, M)$ combination are reported in Table~\ref{tab:grid_success}.

The grid results admit a clear and consistent pattern. First, low MDN capacity ($M{=}1$) fails to capture the multimodal action distributions present in our tasks: the column average for $M{=}1$ is only 11.6\%, with many near-zero entries (e.g., $N_e{=}1,M{=}1=0\%$). This behaviour is expected because a single Gaussian forces regression averaging and cannot represent multiple plausible actions. Second, moderate MDN complexity (particularly $M{=}3$ and $M{=}5$) substantially improves performance: the column averages for $M{=}3$ and $M{=}5$ are 52.2\% and 62.4\% respectively, and the single best cell in the table is $N_e{=}4, M{=}5$ (92\%). This indicates that a moderate number of mixture components is necessary to model the multimodality that remains after semantic conditioning by LCVR.

However, increasing $M$ beyond this moderate range degrades performance: the column averages drop to 48.8\% for $M{=}7$ and collapse to 4.8\% for $M{=}9$. We attribute this to over-parameterization relative to the available data and resulting optimization instability. Too many mixture components increase the dimensionality of the MDN outputs, make component assignment noisier, and can produce ill-conditioned likelihoods (very small variances or unused components), which together amplify training variance and hurt generalization—especially when each expert receives only a limited number of examples.

Regarding expert count, a single expert ($N_e{=}1$) performs poorly (row average 12.8\%) because the monolithic head must represent all task distributions and thus suffers from destructive averaging and gradient conflict. Increasing $N_e$ improves results up to $N_e{=}4$ (row average 49.6\%), which provides the best trade-off between task specialization and sufficient per-expert data for stable MDN fitting. Further increasing to $N_e{=}5$ reduces average performance (42.2\%), likely because each expert then sees fewer examples and the per-expert MDNs are underfitted; additionally, more experts can make gating noisier and increase the difficulty of the auxiliary load-balancing optimization.

Ultimately, the grid search reveals a crucial interaction effect where optimal performance is achieved by balancing task partitioning ($N_e$) with per-expert modeling capacity ($M$). Intuitively, the experts first divide the complex multi-task problem into simpler, specialized sub-problems, allowing each specialist to capture the remaining action ambiguity with only a moderately complex MDN. This principle offers a practical guideline for setups with similar data budgets: a modest number of experts (e.g., 3--4) paired with moderate MDN complexity (e.g., 3--5 components) is most effective. While more data might support larger models, careful monitoring of expert and component utilization would be needed to avoid the instabilities we observed at high $N_e$ or $M$. Based on this analysis, we therefore selected the peak-performing configuration of $N_e{=}4$ and $M{=}5$ as the default for our main comparisons reported in Table~\ref{tab:lmoedp_specialization},and also adopted the same setting in the subsequent ablation study. This finding directly addresses Q2, confirming that the LMoE-DP architecture—with a balanced number of experts and MDN components—effectively mitigates task conflict and captures complex, multimodal action distributions far better than a monolithic policy.

\subsection{Ablation on gradient modulation for training stability}
\begin{figure*}[t]   
  \centering
  \includegraphics[width=\linewidth]{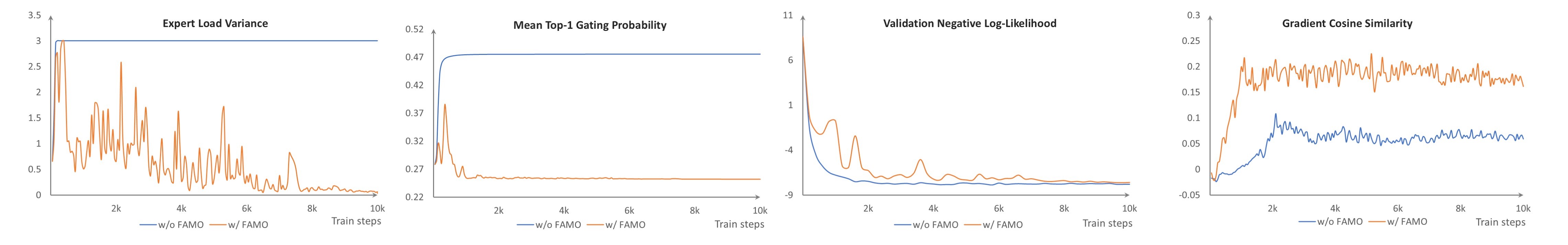}
  \caption{
  Comparison of training dynamics with and without FAMO.  
  FAMO improves gradient alignment, balances expert utilization, and stabilizes the gating distribution,  
  as shown by lower load variance, smoother top-1 probability, reduced Val-NLL, and higher gradient cosine similarity.}
  \label{famo}
\end{figure*}
This experiment aims to quantitatively evaluate the effect of gradient modulation on the training stability of the Mixture-of-Experts (MoE) model.  
We compare two training configurations:  
(1) With FAMO, where gradients from active experts are combined using a minimum-norm projection at each step to reduce inter-task conflicts and balance shared parameter updates;  
(2) Without FAMO, where the same model is trained using standard backpropagation without any gradient modulation.

The experiment uses the same dataset as the grid search in Table~\ref{tab:grid_success}, covering five representative manipulation tasks, each with a canonical variant.  
Both configurations are trained and evaluated under three independent random seeds:10, 20, and 30.  
After training, each model is evaluated in 10 independent trials per task, and all reported results are averaged over the three seeds.

To comprehensively assess the effect of FAMO on both optimization dynamics and final performance, we monitor the following metrics during training and evaluation:

1) Gradient Cosine Similarity. (GradCos)
Measures the alignment among expert gradients on shared parameters, reflecting the level of gradient conflict.

2) Expert Load Variance. (LoadVar)
Quantifies the variance of expert selection frequencies, indicating the balance of expert utilization.

3) Mean Top-1 Gating Probability. (Top1Mean)
Denotes the average maximum gating probability across samples, measuring the sharpness of the routing distribution.

4) Validation Negative Log-Likelihood (Val-NLL).  
Evaluates model generalization by measuring the validation negative log-likelihood.

5) Real-World Task Success Rate.  
Measures the average success rate across five real-robot tasks.

For the training metrics—GradCos, LoadVar, Top1Mean, and Val-NLL, we first compute the mean over time across all training steps. We then average these values over three random seeds. For the real-world evaluations, each task is performed 10 times under each seed, with the success rate defined as the ratio of successful trials to 10. These success rates are then averaged across the three seeds. The final table summarizes the average task success rates under both settings, with and without FAMO.

\begin{table}[htbp]
\centering
\caption{Task success rates (\%) with and without FAMO.}
\label{tab:famo_success}
\begin{tabular*}{\dimexpr0.8\columnwidth\relax}{@{\extracolsep{\fill}}lcc}
\toprule
\textbf{Task} & No FAMO & \textbf{FAMO} \\
\midrule
Stack Cups & 30.0 & \textbf{76.6}\\
Put Fruit in Plate & 73.3 & \textbf{93.3}\\
Hang Mug & 56.6 & \textbf{90.0}\\
Put Item in Barrel & 50.0 & \textbf{86.6}\\
Put Tennis in Mug & 63.3 & \textbf{83.3}\\
\midrule
Average Success Rate & 54.6 & \textbf{85.9}\\
\bottomrule
\end{tabular*}
\end{table}
As shown in Fig.~\ref{famo}, gradient modulation with FAMO produces clear improvements across all four training metrics. Without FAMO, the expert load variance remains saturated at a high level, indicating severe imbalance where only a few experts dominate the routing. In contrast, FAMO rapidly reduces the variance toward zero, ensuring balanced utilization of all experts. Similarly, the mean top-1 gating probability without FAMO quickly converges to $\approx$0.47, reflecting near-deterministic routing to a single expert, whereas with FAMO it stabilizes around $1/N_e$ (0.25 for $N_e{=}4$), demonstrating healthy expert competition. For validation NLL, both configurations converge to similar values, but FAMO yields smoother trajectories with reduced instability. Finally, gradient cosine similarity is consistently higher with FAMO (0.15--0.22) compared to the much lower values observed without FAMO (0.05--0.08), indicating better gradient alignment and effective mitigation of conflicts among experts.

Beyond these training dynamics, Table~\ref{tab:famo_success} highlights the practical impact of FAMO on real-robot performance. Across five representative manipulation tasks, the average success rate increases from 54.6\% without FAMO to 85.9\% with FAMO, with particularly large improvements on complex tasks such as ``Stack Cups'' (from 30\% to 76.6\%) and ``Put Item in Barrel'' (from 50\% to 86.6\%). These results confirm that stabilizing gradient updates not only improves expert specialization during training but also translates into tangible improvements in real-world task execution. Taken together, this ablation directly addresses Q3, demonstrating that integrating gradient modulation is crucial for achieving both stable optimization of shared components and high final task success rates in multi-task robotic manipulation.

\subsection{Model parameters and inference latency}
Table~\ref{tab:latency_params} summarizes the model size and inference latency. While LMoE-DP has a larger total parameter count (208.09M), most of it comes from the frozen visual backbone, with only 58.47M parameters being trainable. The average inference latency is 96.32\,ms per action on an RTX 4060 Ti, which supports real-time control above 10\,Hz. This design allows the model to achieve strong multi-task performance while remaining lightweight and efficient for deployment.   

\begin{table}[htbp]
\centering
\caption{Model parameters and inference latency.}
\label{tab:latency_params}
\renewcommand{\arraystretch}{1.3}
\setlength{\tabcolsep}{2pt} 
\begin{tabular}{lccc}
\toprule
\textbf{Method} & Total Params (M) & Trainable Params (M) & Latency (ms) \\
\midrule
ACT~\cite{ACT} & 83.88 & 83.88 & 7.44 \\
DP~\cite{DP}   & 55.32 & 55.32 & 63.78 \\
LMoE-DP (Ours) & 208.09 & 58.47 & 96.32 \\
\bottomrule
\end{tabular}
\end{table}

\section{Conclusion}
We present an end-to-end framework for multi-task robotic manipulation integrating LCVR and LMoE-DP. LCVR enhances visual representations with semantic grounding to mitigate perceptual ambiguity, while LMoE-DP uses a sparse MoE structure with gradient modulation for task specialization and stable optimization. Together, they enable efficient multi-task imitation learning.

Despite its effectiveness, several limitations remain. The reliance on a single fixed-view RGB camera makes the system sensitive to occlusion and limits depth perception. Moreover, the current language conditioning operates at the static instruction level, without dynamic adaptation to task progress or environmental feedback. Future work will explore multi-view or depth-enhanced perception, temporal and interactive language grounding, and adaptive expert learning mechanisms to further improve generalization and robustness in unstructured real-world environments. 

\bibliographystyle{IEEEtran}
\bibliography{reference} 
\end{document}